\documentclass[letterpaper, 10 pt, conference]{ieeeconf}  

\usepackage{graphicx}
\usepackage{enumerate}
\usepackage{amsmath}
\usepackage{algorithm}
\usepackage[noend]{algpseudocode}
\usepackage{subfigure}
\usepackage{chngcntr}
\usepackage{graphicx}
\usepackage{bm}
\usepackage{gensymb}
\usepackage{tabularx}
\makeatletter
\let\NAT@parse\undefined
\makeatother
\usepackage[table,xcdraw]{xcolor}
\usepackage{graphicx}

\usepackage[sort&compress,numbers]{natbib}

\usepackage{url}
\usepackage{tikz}

\usepackage{caption}
\usepackage{hyperref}
\hypersetup{
	colorlinks = true
}

\usepackage[colorinlistoftodos]{todonotes}

\usepackage[font=footnotesize,labelfont=bf,tableposition=top]{caption}

\usepackage{todonotes}

\usepackage{etoolbox}
\makeatletter
\patchcmd{\@makecaption}
{\scshape}
{}
{}
{}
\makeatletter
\patchcmd{\@makecaption}
{\\}
{.\ }
{}
{}
\makeatother

\usepackage{color}

\IEEEoverridecommandlockouts

\usepackage{caption}
\usepackage[singlelinecheck=false, font=footnotesize]{caption}

\usepackage{etoolbox}
\makeatletter
\patchcmd{\@makecaption}
{\scshape}
{}
{}
{}
\makeatletter
\patchcmd{\@makecaption}
{\\}
{.\ }
{}
{}
\makeatother

\usepackage{amsfonts}
\usepackage{amssymb}

\makeatletter
\def\BState{\State\hskip-\ALG@thistlm}
\makeatother

\usepackage{epstopdf}
\usepackage{epsfig}
\usepackage{multirow}
\usepackage{soul}

\begin{document}
	\title{\Large \bf SegICP-DSR: Dense Semantic Scene Reconstruction and Registration}
	
	\author{Jay M. Wong, Syler Wagner, Connor Lawson${}^\dagger$, Vincent Kee${}^\ddagger$, Mitchell Hebert, Justin Rooney, \\Gian-Luca Mariottini, Rebecca Russell, Abraham Schneider, Rahul Chipalkatty, and David M.S. Johnson\\
		\emph{Draper, Cambridge, MA, USA }  \\ 
		\thanks{Author associated with the Georgia Institute of Technology${}^\dagger$ and the Massachusetts Institute of Technology${}^\ddagger$. We thank Tiffany Le and Lei Hamilton for their initial contributions as well as Leslie Pack Kaelbling, Tom{\'a}s Lozano-P{\'e}rez, Scott Kuindersma, and Russ Tedrake for their insight and feedback. 
			\textbf{Corresponding author:} Jay M. Wong {\href{mailto:jmwong@draper.com}{\texttt{jmwong@draper.com}}}}
	}
	\maketitle

	\IEEEpeerreviewmaketitle
	
	\begin{abstract}
		
		%
		
		To enable autonomous robotic manipulation in unstructured environments, we present \emph{SegICP-DSR}, a real-time, dense, semantic scene reconstruction and pose estimation algorithm that achieves $mm$-level pose accuracy and standard deviation ($7.9 \,mm, \sigma\mathord{=}7.6 \,mm$ and $1.7\deg, \sigma\mathord{=}0.7\deg$) and successfully identified the object pose in $97\%$ of test cases. This represents a $29\%$ increase in accuracy, and a $14\%$  increase in success rate compared to \emph{SegICP} in cluttered, unstructured environments. The performance increase of \emph{SegICP-DSR} arises from (1) improved deep semantic segmentation under adversarial training, (2) precise automated calibration of the camera intrinsic and extrinsic parameters, (3) viewpoint specific ray-casting of the model geometry, and (4) dense semantic \emph{ElasticFusion} point clouds for registration. We benchmark the performance of \emph{SegICP-DSR} on thousands of pose-annotated video frames and demonstrate its accuracy and efficacy on two tight tolerance grasping and insertion tasks using a KUKA LBR iiwa robotic arm.

	\end{abstract}
	
	\section{Introduction}
	Robotic manipulation in unstructured environments is an unsolved research problem \cite{pratt2013darpa, wurman2015amazonpicking}, and recent work has led to the development of systems capable of perceiving the world and extracting task-relevant information (e.g. object identity and pose) useful for manipulation and planning \cite{fallon2015architecture,jonschkowski2016probabilistic,Zeng2017multi, Wong2017segicp}. As a result, today's state-of-the-art systems are making substantial progress towards coarse manipulation (e.g. picking and sorting) but still struggle with manipulation tasks that  demand tight tolerance (${\ll}\,1\,cm$) in unstructured environments. 
	
	
	
	
	
\begin{figure}
		\centering
		\vspace{7.5pt}
		\includegraphics[width=0.485\textwidth]{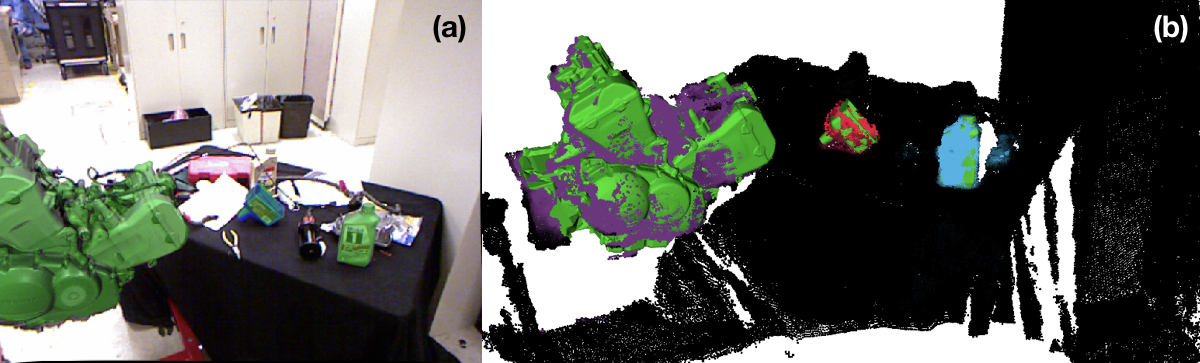}
		\caption{\textbf{
				SegICP-DSR}: (a) object CAD models (shown in green) projected back into the camera frame using estimated \emph{SegICP-DSR} object poses were obtained by (b) model-to-scene registration to a dense point cloud generated by \emph{ElasticFusion} after fusing a trajectory of semantically labeled depth images to create a dense, labeled reconstruction of the scene.}
		\label{fig:pipeline}
	\end{figure}

	To address this, we present an object-recognition and pose-estimation algorithm --- \emph{SegICP-DSR}, that works by (a) fusing a series of RGB-D observations into a dense, semantically-labeled point cloud, then (b) generating candidate point clouds from existing CAD models by ray-casting, and finally (c) performing model-to-scene registration via iterative closest point (ICP) registration using a model-to-scene correspondence metric \cite{Wong2017segicp} (see Fig. \ref{fig:pipeline} and \ref{fig:arch}). \emph{SegICP-DSR} provides high accuracy, low variance pose estimates in unstructured environments, because it attenuates sensor noise by smoothing successive measurements and handles dynamic scenes where obstacles may become occluded due to object or camera motion. 
	
	
	In our test-cases, \emph{SegICP-DSR} achieves $mm$-level position accuracy and standard deviation ($7.9 \,mm, \sigma\mathord{=}7.6 \,mm$) and degree-level orientation accuracy and standard deviation ($1.7\deg, \sigma\mathord{=}0.7\deg$), which enables tight tolerance manipulation in unstructured environments. 
	
	\section{Related Work}
	
	Accurate pose estimation in cluttered, unstructured environments is challenging, and many successful manipulation techniques (e.g. visual servoing) bypass the problem entirely. Several recent manipulation approaches have shown great success by avoiding an explicit representation of object pose and directly mapping raw sensor observations to motor behavior \cite{Levine2016end,Mahler2016dexnet,Mahler2017dexnet}. However, these approaches do not provide the semantic expressiveness required for symbolic task planners \cite{Kaelbling2013}, which are convenient for complex, long-horizon, multi-step manipulation tasks. Furthermore, object recognition and pose estimation provide an elegant, compact representation of  high-dimensional sensor input. In this regard, we focus primarily on approaches that first solve the object-identification and pose-estimation problem and subsequently tackle manipulation. 
	
	
	\subsection{Point Cloud Registration}
	
	Recent efforts in pose estimation have contributed to faster and more precise architectures for locating objects of interest in cluttered scenes. Among these various approaches are point cloud registration algorithms \cite{besl1992method,rusinkiewicz2001efficient,yuan20163d,schmidt2014dart,Zeng2017multi, Wong2017segicp}, which are prone to converge to local minima and are brittle to partial or full occlusions. Global registration schemes have been proposed, but are not yet real-time, do not scale \cite{izattglobally}, or require additional information (e.g. surface normals) \cite{zhou2016fast} which can degrade the solution at specific perspectives. To counter these shortcomings, several solutions have been proposed which first segment the scene point cloud, then use a local registration method (usually ICP) for model-to-scene alignment with either a prior pose estimate or the segmentation centroid as a seed \cite{Zeng2017multi, Wong2017segicp}. These approaches also fuse multiple observations from different viewpoints to improve the pose estimate and can run in real-time (${\sim}{15}~\text{Hz}$), but their pose estimation errors still plateau at ${\sim}{1}\,cm$, which is insufficient for tight-tolerance robotic manipulation tasks such as an open-loop peg-in-hole insertion (note that closed-loop manipulation strategies employing visual or force feedback may also be used to further improve performance). For this reason, we investigate combining semantic segmentation with dense scene reconstruction to further reduce pose estimation errors and variance by increasing the density and accuracy of the semantically-labeled point clouds used for registration. 
	\begin{figure}
		\centering
		\includegraphics[width=0.475\textwidth]{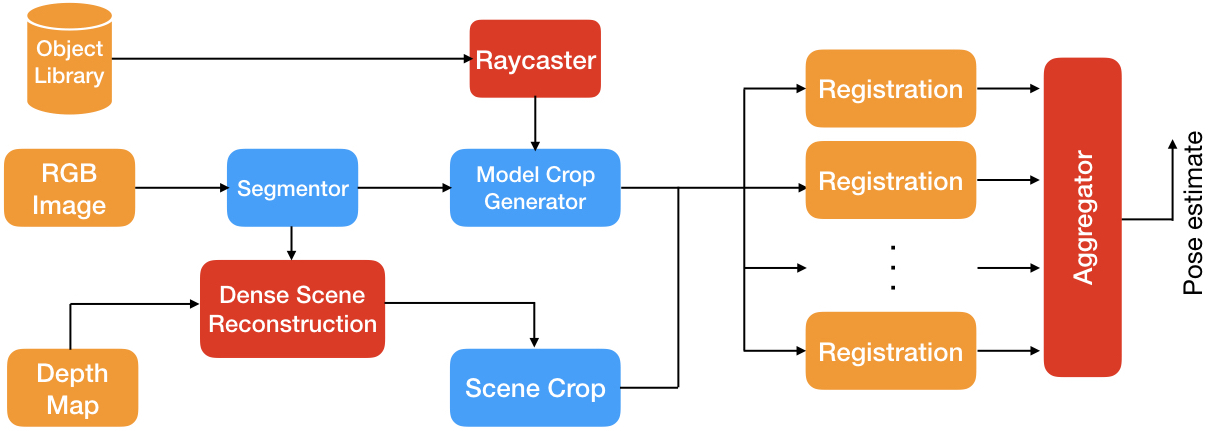}
		\caption{\textbf{SegICP-DSR architecture:} extending \emph{SegICP} new modules (red) and updated modules (blue) are introduced primarily in support of dense point cloud registration, reducing errors and variance in pose estimation.}
		\label{fig:arch}
	\end{figure}

	\subsection{Dense Scene Reconstruction}
	
	A number of real-time dense visual SLAM (simultaneous localization and mapping) systems have been developed \cite{seitz1999photorealistic,newcombe2010live,geiger2011stereoscan,newcombe2011kinectfusion,whelan2015real,stuckler2014multi,zhou2013dense,Whelan2015elastic,Whelan2016elastic}. While many of these approaches scale well, they have poor performance with locally loopy trajectories \cite{whelan2015real}, perform no explicit map reconstruction, or require pose graph optimization steps and merging key surface element (surfel) views to create the final map \cite{stuckler2014multi}. Systems that place large emphasis on the accuracy of the reconstructed map over estimated trajectories are offline \cite{zhou2013dense}. Thus, for our application, we elect to begin with \emph{ElasticFusion} \cite{Whelan2015elastic,Whelan2016elastic} because of its real-time operation and map-centric approach, which avoids post-processing or pose graph optimization and aims to use surface prediction at every frame for simultaneous dense mapping and camera pose estimation. \emph{ElasticFusion} has found numerous recent applications, such as \emph{SemanticFusion}, a system that produces semantically fused dense reconstruction \cite{mccormac2017semanticfusion}. \emph{SemanticFusion's} architecture leveraged a convolutional neural network for segmentation, \emph{ElasticFusion} for visual SLAM and frame-to-frame correspondences, and a Bayesian update scheme to track object class probabilities per surfel. In a similar approach, \emph{DA-RNN} \cite{xiang2017rnn}, a 3D semantic mapping system, has demonstrated high performance using \emph{KinectFusion} \cite{newcombe2011kinectfusion} to produce 3D semantic scenes. 
	
	Additionally, \emph{LabelFusion} has demonstrated the use of dense scene reconstruction to generate large datasets of labeled segmentation and pose images---specifically a million object instances in multi-object scenes \cite{Marion2017}. Contrary to this approach, where a hand-aligned object pose is used to generate segmentation, we are interested in the reverse problem: high accuracy object pose estimation using online segmentation. 
	
	\section{Technical Approach}
	
	We present a novel architecture (shown in Figure~\ref{fig:arch}) that improves upon \emph{SegICP} \cite{Wong2017segicp}---explicitly tackling the following shortcomings: (1) sub-par RGB-D calibration and (2) segmentation errors,  both of which result in pose-estimation errors due to incorrect points being included in the local registration step, (3) sensor noise which leads to large variance in the pose estimates, and (4) registration errors caused by model-crops which include points not present in the observed point cloud.
	
	To address the sub-par RGB-D calibration, we developed an automated method to simultaneously calibrate the extrinsic and intrinsic parameters of our color and depth cameras, which resulted in a $1.5$ pixel residual error (Section~\ref{sec:calib}). Next, we adversarially trained a semantic segmentation convolutional neural net (CNN) to further increase segmentation accuracy (Section~\ref{sec:segmentation}). Exploiting this precise calibration and segmentation performance, we replaced the pixel-intensity-based photometric error term in \emph{ElasticFusion} \cite{Whelan2015elastic} with the semantic label, resulting in a real-time, labeled, dense, scene reconstruction (Section~\ref{sec:fuse}). Using realistic model point clouds generated by ray-casting (Section~\ref{sec:ray-cast}), we then performed model-to-scene registration to acquire each object's pose estimate, while leveraging a time-history of viewpoints for more accurate and lower variance pose estimation.

	\begin{figure}
		\centering
		\includegraphics[width=0.475\textwidth]{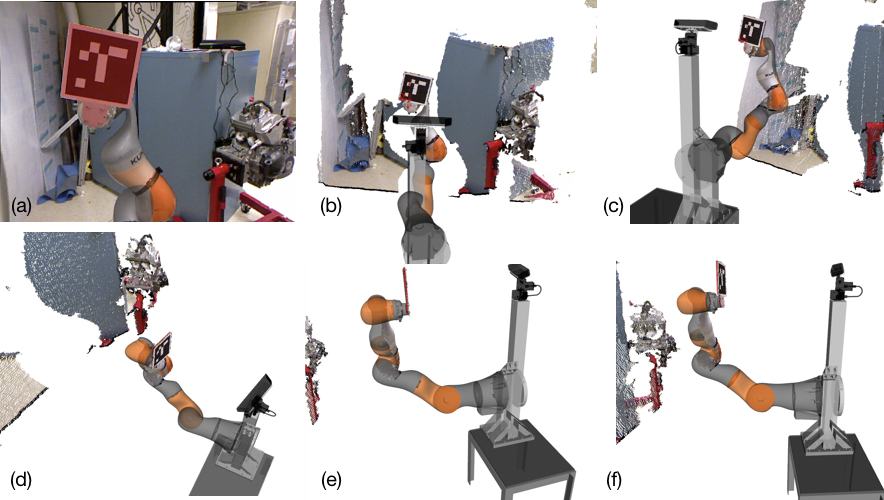}
		\caption{\textbf{Calibration result:} (a) projected robot model overlaid onto the Kinect1 image has under $1.5$ pixel error. (b-f) illustrates various 3D viewpoints; note the point cloud alignment with the robot's mesh (both at the calibration target in red and along the arm's surface).}
		\label{fig:calibration}
	\end{figure}
	
	\subsection{Simultaneous Intrinsic and Extrinsic Calibration}
	\label{sec:calib}
	\begin{figure*}
		\centering
		\includegraphics[width=\textwidth]{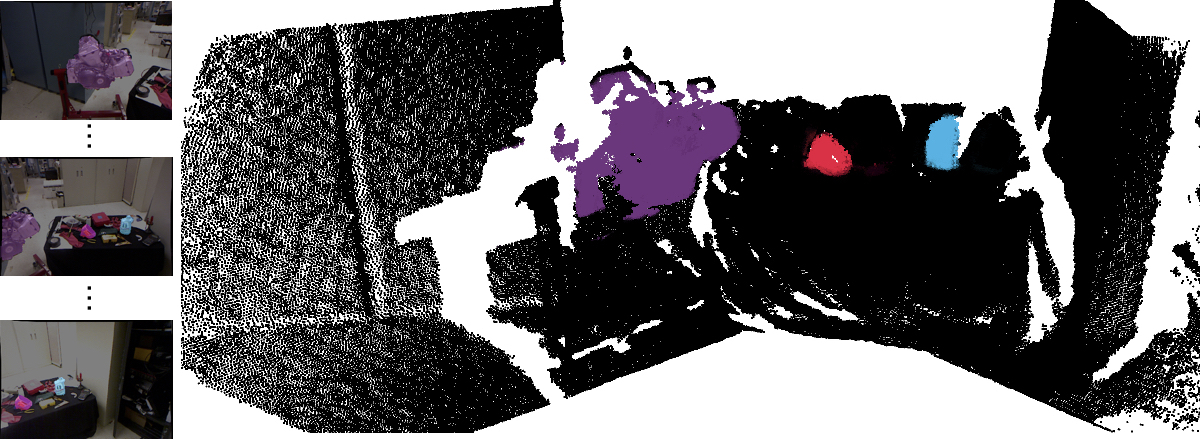}
		\caption{\textbf{An example of a dense labeled reconstructed scene point cloud} with $207,535$ points generated using the trajectory of segmentations (left). The objects of interest are classified as purple corresponding to engine, red to funnel, blue to oil bottle and black for background. The underlying camera images and their segmentation overlays are illustrated to the left. Notice that all clutter on the table are completely ignored and treated as background in the dense reconstruction.}			
		\label{fig:fused-scene}	
	\end{figure*}			
	
	Our automated data collection and optimization procedure to calibrate the Kinect1 RGB and IR cameras works as follows. First we collect a set of $N$ measured observation pairs, consisting of the robot arm end effector position computed from forward  kinematics and the four corner pixel locations of the calibration target as measured by an Aruco tag detector \cite{aruco2014}.  Next we fit this data with a first order Brown-Conrady distortion \cite{Brown71close} pinhole model using L-BFGS-B \cite{zhu1997algorithm} with bounds centered around the initial parameter estimates for each camera by optimizing for projection and distortion parameters that minimize pixel-wise projection error in image space. The minimization is given by:
	\begin{equation}
	\min_{A, d, f, c}{\sum_{i=1}^{N}{\|x_i - \hat{x}_i\|^2}}
	\label{eqn:calib_min}
	\end{equation}
	where $A \in SE(3)$ is the camera extrinsic matrix; $d = [k_1, k_2, p_1, p_2] \in \mathbb{R}^4$ is the vector of radial and tangential distortion coefficients; $f, c \in \mathbb{R}^2$ are the focal lengths and principal points, respectively; and $\hat{x}_i \in \mathbb{R}^2$ is the projection of $X_i$ (the position of the target corners as measured by forward kinematics) into the unrectified image using the model defined by the minimization parameters. After applying this procedure, our Kinect1 camera achieved a $1.5$ average pixel error over a calibration dataset of $480$ RGB and IR images (the results of which are shown in Fig.~\ref{fig:calibration}).

	\subsection{Segmentation via Adversarial Networks}
	\label{sec:segmentation}
	Segmentation plays an important role in our overall framework.  It crops the scene point cloud to obtain only the points corresponding to the objects of interest, which reduces the number of local minima available to ICP. Thus, as prior results indicate, our pose accuracy is directly related to the quality of segmentation \cite{Wong2017segicp}. To improve our segmentation performance (measured as intersection over union or IOU), we investigated various state-of-the-art network architectures and trained them  adversarially \cite{Goodfellow2014generative}. Previous results indicate that training two networks (one as a semantic segmentor and another as an adversarial discriminator) leads to increased labelling accuracy \cite{luc2016semantic}.
	
	Table~\ref{table:networks} outlines the network architectures we tested for better segmentation quality. These networks were all trained and evaluated using PyTorch\footnote{PyTorch is a tensors and dynamic neural networks Python package with strong GPU acceleration \href{http://pytorch.org}{\texttt{http://pytorch.org}}.} with a dataset of $17084$ multi-object images in the automotive maintenance domain. %
	\vspace{-5pt}
	\begin{table}[H]
		\centering
		\begin{tabular}{|l|c|c|c|}
			\hline
			& \cellcolor[HTML]{EFEFEF}\textbf{IOU Metric} & \cellcolor[HTML]{EFEFEF}\textbf{Parameters} & \cellcolor[HTML]{EFEFEF}\textbf{Forward pass*} \\ \hline
			\cellcolor[HTML]{EFEFEF}\textbf{SegNet-A}  & $0.901, \sigma\mathord{=}0.081$               & $29$M                 &  $65\,ms$ / $25\,ms$                               \\ \hline
			\cellcolor[HTML]{EFEFEF}\textbf{DeepLab} \cite{chen2016deeplab}                           & $0.872, \sigma\mathord{=}0.057$       & $43$M                 & $175\,ms$ / $70\,ms$                          \\ \hline
			\cellcolor[HTML]{EFEFEF}\textbf{SegNet} \cite{badrinarayanan2015segnet}                            & $0.850, \sigma\mathord{=}0.159$               & $29$M                 &  $65\,ms$ / $25\,ms$                         \\ \hline
		\end{tabular}
		\vspace{5pt}
		\caption{\textbf{Various segmentation architectures} considered for \emph{SegICP-DSR}. The -A networks were trained adversarially. *The forward pass evaluations are averages reported with input image sizes of $640\times480$ and $320\times320$ evaluated on a nVidia TitanXp GPU.}
		\label{table:networks}
	\end{table}			
	\vspace{-10pt}
	\noindent We found that training segmentors adversarially increased the segmentation quality, \emph{even between networks with identical architectures}---which is in agreement with \cite{luc2016semantic}. As such, for \emph{SegICP-DSR}, we use an adversarially trained encoder-decoder \emph{SegNet-A} architecture for segmentation. 
	
	\subsection{Fusing Consistent Segmentation Frames}	
	\label{sec:fuse}
	\emph{ElasticFusion} uses trajectories of raw RGB-D frames as input and provides an estimated camera pose in addition to a dense reconstruction of the scene (by minimizing the cost $E_{rgb}$ which enforces constant intensity at each pixel based on the sum of the RGB values). Our goal is to instead create a labeled, dense, unorganized point cloud to be used for efficient segmentation and registration (as shown in Figure~\ref{fig:fused-scene}). The \emph{SemanticFusion} \cite{mccormac2017semanticfusion} system does by using  raw RGB-D camera frames as input to \emph{ElasticFusion} and relying on a Bayesian update scheme to generate the final map with the output of \emph{ElasticFusion} and a segmentor. Instead, we directly provide \emph{ElasticFusion} with a labeled point cloud, rather than a colorized (RGB) point cloud, avoiding the Bayesian update step.
	
	Therefore, \emph{SegICP-DSR} aims to find the motion parameters $\bm \xi$ to minimize the difference between the latest segmentation label for each measured point $\mathcal{S}(\mathbf u, \mathcal{C}^l_t)$ and the label of the transformed active model of the last frame $\mathcal{S}(\zeta(\bm \xi) ,\hat{\mathcal{C}}^a_{t-1})$, where $\zeta(\bm \xi)$ is the transformed point corresponding to $\mathbf u$ from the active model. For simplicity in the formulation, we substituted $\zeta(\bm \xi) = \bm\pi(\mathbf K\exp(\hat{\bm{\xi}})\mathbf T\mathbf p(\mathbf u,\mathcal{D}^l_t))$ (see \cite{Whelan2015elastic} for details).
	
	We represent this cost over the label differences as $E_{label}$, which we use instead of the existing photometric error $E_{rgb}$ in our \emph{ElasticFusion} implementation \cite{Whelan2015elastic}:
	\begin{equation}
	E_{rgb} \rightarrow E_{label} = \sum_{u\in \Omega}\big( \mathcal{S}(\mathbf u, \mathcal{C}^l_t) - \mathcal{S}(\zeta(\bm\xi) ,\hat{\mathcal{C}}^a_{t-1} ) \mathord{\big)}^2
	\label{eqn:efusion_min}
	\end{equation}
	where the term $\mathcal{S}:\mathbf{u}, \mathcal{C} \mapsto label$ uses the output of \emph{SegNet} to assign integer values to pixels according to their semantic label. 
	The $E_{label}$ cost encourages the reconstruction algorithm to provide correspondences between frames directly in the space of segmentation labels given by $\mathcal{S}$, instead of intensity. As such, this allows us to bypass the Bayesian update scheme and directly take the raw output of \emph{ElasticFusion} as our semantically-labeled point cloud without an extra update step. 
	
	\textbf{Scene Cropping.} In order to obtain $n$ object cropped scene clouds for registration, we iterate through all points in the reconstructed point cloud and select the semantic labels corresponding to each of the objects, which is an efficient $O(N)$ operation. The dense reconstruction output typically has fewer points than a single-frame Kinect1 point cloud. The Kinect1 point cloud at $640\times480$ resolution has $307,200$ points whereas typical dense scene reconstructions contain  $150,000$ to $300,000$ points (see Fig.~\ref{fig:fused-scene}). This means that the per-object scene cropping is  \emph{faster} on the dense reconstruction than the raw sensor point cloud.			
	\begin{figure}
		\centering
		%
		\includegraphics[width=0.485\textwidth]{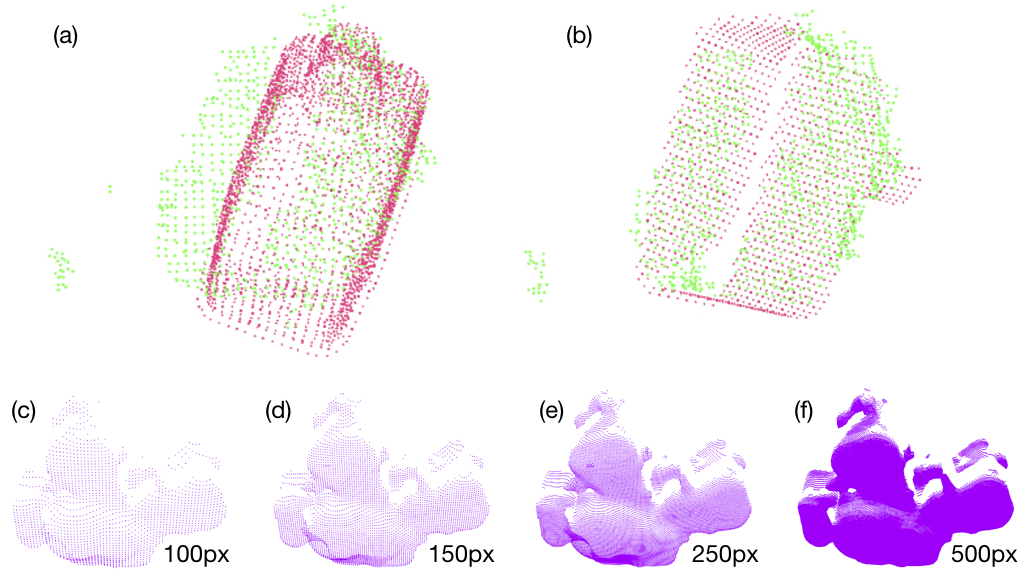}
		\caption{\textbf{Ray-casting model crops:} (a) Extraneous model points cause local minima in registration for the funnel leading to a poor pose-estimate, while  (b) ray-casting renders a realistic candidate model cloud and results in successful registration. Model points are illustrated in red, while the scene points are indicated in green. (c-f) Examples of engine model crop candidates with varying point density.}
		\label{fig:ray-cast-error}
	\end{figure}

	\subsection{Generating Realistic Candidate Point Clouds}
	\label{sec:ray-cast}
	Point-to-point registration of observed point clouds to points generated from a CAD model is challenging because the ICP optimization contains many local minima. Many of these minima arise if  extraneous points are included in the  generated model point cloud, which do not have corresponding points in the observed point cloud due to occlusions or the camera viewpoint. If the ICP least-squares minimization finds one of these minima, it results in registration errors as seen in Figure~\ref{fig:ray-cast-error}. Because of this, we can not use a trivial model-cropping strategy to generate these point clouds; instead, we use a VTK implementation of ray-casting \cite{schroeder2004visualization} to render plausible model point clouds,  providing sets of visible points for the object at various azimuths and elevations around the object mesh (see Fig. \ref{fig:ray-cast-error}). This approach is robust to objects with concave components (e.g. funnels) and removes many of the undesirable local minima in the registration step. In addition, ray-casting allows us to conveniently control the point density of individual object clouds, which also contributes to better registration solutions. 
	
	
	Additionally, by dynamically ignoring candidate model crops based on their euclidean and quaternion metric distance, we achieve a $2\times$--$3\times$ registration speedup by reducing the number of ICP registrations which must be performed at each time-step.

	
	\section{Experiments}	
	\begin{figure}
		\centering
		\subfigure{\includegraphics[height=2.5cm]{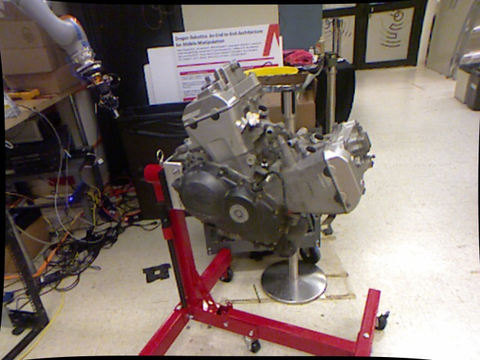}}
		\subfigure{\includegraphics[height=2.5cm]{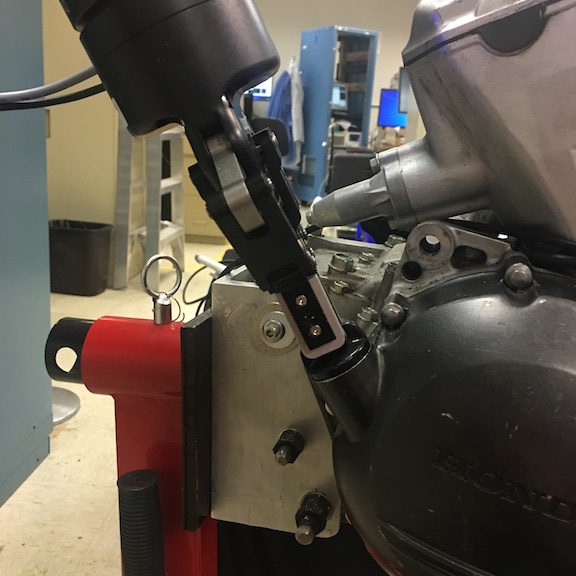}}
		\subfigure{\includegraphics[height=2.5cm]{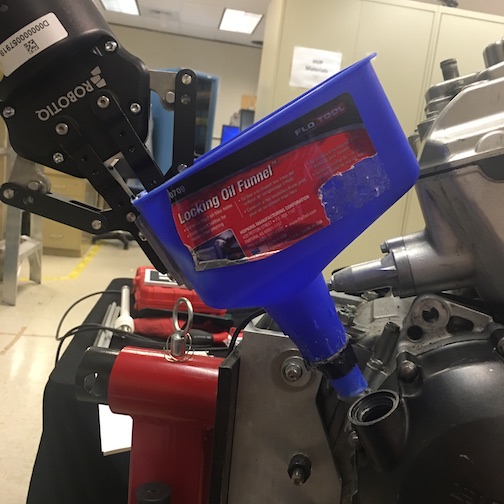}}
		\vspace{-10.5pt}
		\caption{\textbf{Experiment methodologies:} Our experiments consists of (1) a $4493$ frame RGB-D \emph{LabelFusion} benchmarking dataset, (2) tight-tolerance engine cap grasping, and (3) tight-tolerance funnel insertion (L-R)}
		\label{fig:exp-pic}
	\end{figure}%
	We performed three experiments to evaluate the performance of \emph{SegICP-DSR} as compared to \emph{SegICP} \cite{Wong2017segicp} on real-world, tight-tolerance manipulation tasks, all on physical hardware: either on a calibrated Kinect1 RGB-D sensor or on our integrated KUKA robotic system (a KUKA LBR iiwa14 arm mounted onto a torso and equipped with a Robotiq 2F-85 Adaptive Gripper, DirectPerception PTU-E46 head, and Microsoft Kinect1 RGB-D sensor). 
	
	
	These experiments (illustrations in Figure~\ref{fig:exp-pic}) are aimed at evaluating both perception accuracy and variance as well as our ability to perform  tight-tolerance manipulation in the real  world. They are as follows:
	
	\begin{figure*}
		\begin{table}[H]
			\hspace{45pt}%
			\begin{tabular}{|l|c|}
				\hline 
				\multicolumn{2}{|c|}{	\emph{SegICP} \cite{Wong2017segicp}}  \\ \hline
				\cellcolor[HTML]{EFEFEF}Success ($\%$)& $82.4\,\%$               \\ \hline
				\cellcolor[HTML]{EFEFEF}Translation norm ($mm$)& $11.6\,mm, \sigma\mathord{=}14.8\,mm$               \\ \hline
				\cellcolor[HTML]{EFEFEF}Quaternion metric ($deg$)& $1.61\,\deg, \sigma\mathord{=}2.35\,\deg$               \\ \hline
			\end{tabular}
		\end{table} 
		\vspace{-12.5pt}
		\subfigure{\includegraphics[width=0.5\textwidth]{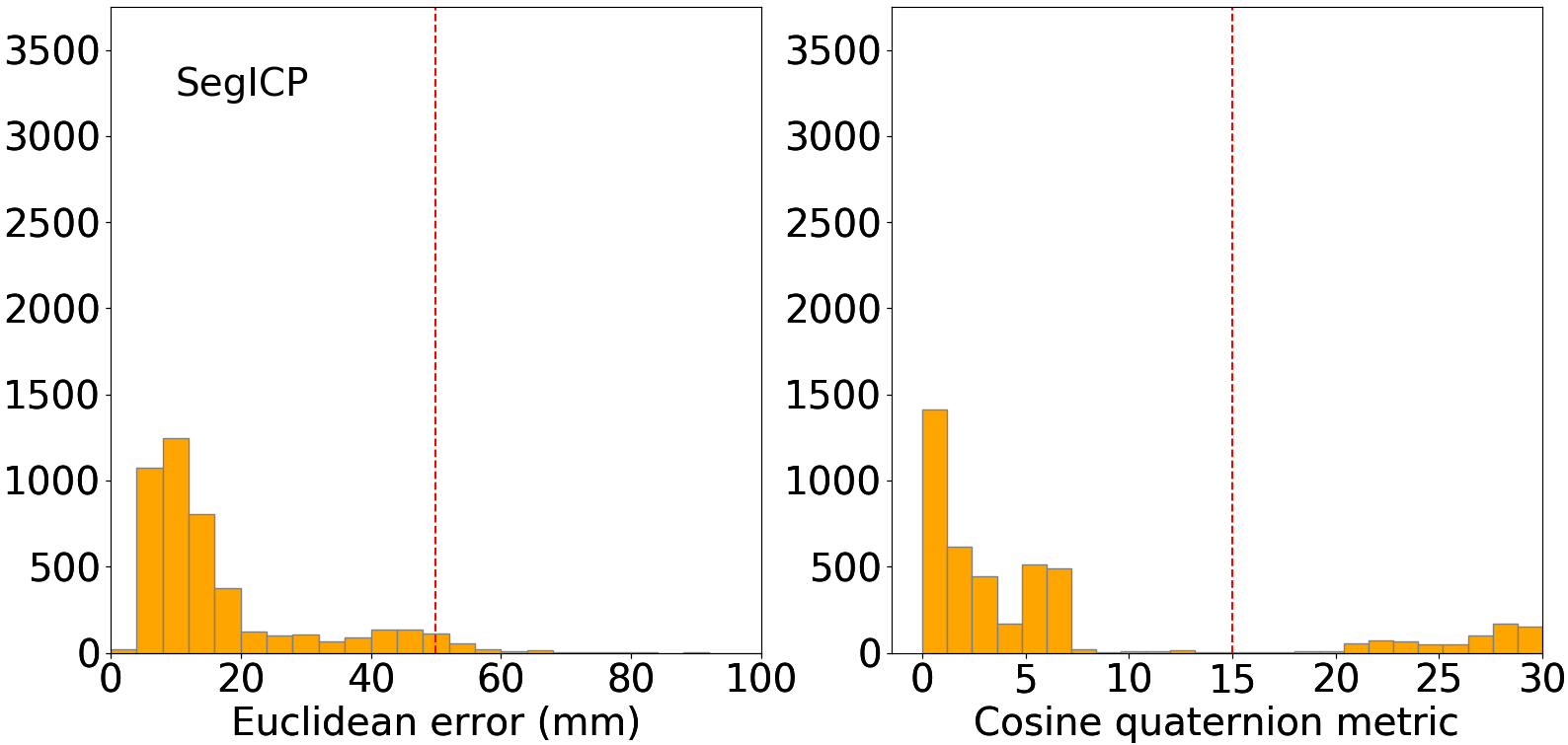}}%
		\vspace{-180pt}
		\begin{table}[H]
			\hspace{295pt}%
			\begin{tabular}{|l|c|}
				\hline 
				\multicolumn{2}{|c|}{	\emph{SegICP-DSR}}  \\ \hline
				\cellcolor[HTML]{EFEFEF}Success ($\%$)& $96.7\,\%$               \\ \hline
				\cellcolor[HTML]{EFEFEF}Translation norm ($mm$)& $7.9\,mm, \sigma\mathord{=}7.6\,mm$               \\ \hline
				\cellcolor[HTML]{EFEFEF}Quaternion metric ($deg$)& $ 1.70\,\deg, \sigma\mathord{=}0.71\,\deg$               \\ \hline
			\end{tabular}
		\end{table} 
		\vspace{-12.5pt}
		\hspace{252.5pt}%
		\subfigure{\includegraphics[width=0.5\textwidth]{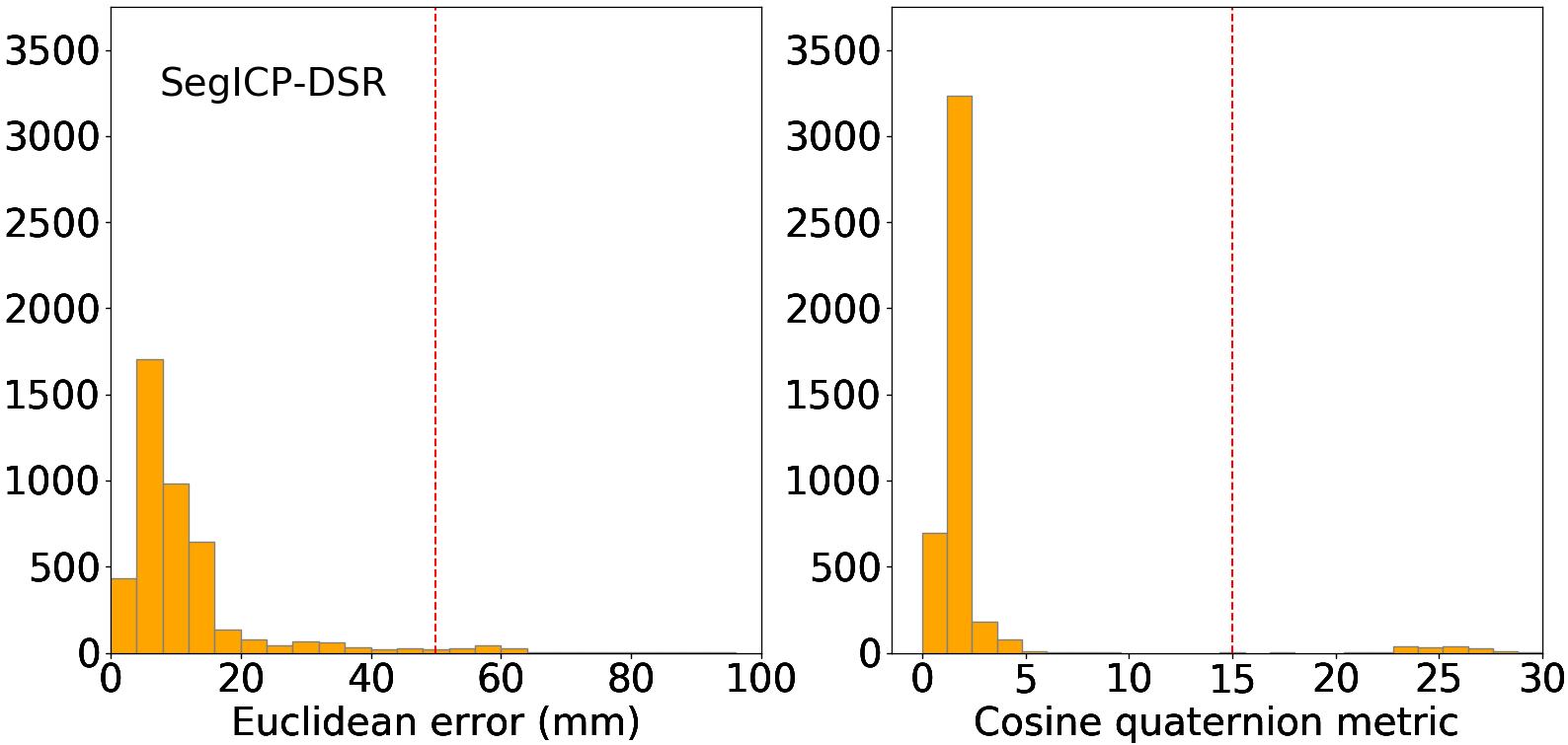}}%
		\caption{\textbf{Experiment 1: \emph{LabelFusion} dataset benchmarking.} Of the $4493$ video frames, \emph{SegICP-DSR} achieves notably higher success measures (with metric $<0.05 m$ and $< 15\deg$ \cite{Wong2017segicp}). Furthermore, of the successful frames, \emph{SegICP-DSR} attains more accurate pose estimations on average, with considerably lower standard deviation.	}
		\label{fig:labelfusion-dataset}
	\end{figure*}		
	
	\vspace{5pt}
	\textbf{-- Experiment 1: LabelFusion dataset.} Benchmarking accuracy and variance of \emph{SegICP-DSR} vs. \emph{SegICP} on a $4493$ frame video generated by LabelFusion\footnote{LabelFusion is a pipeline to rapidly generate annotated segmentation and pose RGB-D datasets \href{http://labelfusion.csail.mit.edu}{\texttt{http://labelfusion.csail.mit.edu}}} \cite{Marion2017}. 
	
	\vspace{5pt}
	\textbf{-- Experiment 2: Tight-tolerance grasping.} Sub-$cm$ level accuracy is required for successfully grasping the oil-filler cap whose radius is $17\,mm$ (with gripper width $22\,mm$). 
	
	\vspace{5pt}
	\textbf{-- Experiment 3: Tight-tolerance, semi-rigid peg-in-hole insertion.} This is a sub cm-level precision insertion task with a funnel tip radius of $8\,mm$ and engine hole radius of $11.5\,mm$.
	
	\subsection*{Experiment Results}
	We perform the following experiments in order to confirm the accuracy and stability of our proposed \emph{SegICP-DSR} approach, integrate it in the loop with model-based manipulation, and explore its limitations in real-world manipulation tasks. The results of these experiments are as follows:
	
	\vspace{5pt}
	\textbf{-- Experiment 1: \emph{LabelFusion} dataset.} In order to generate a large dataset for evaluation, we used \emph{LabelFusion} \cite{Marion2017} to create various multi-frame videos with a motorcycle engine in the scene. \emph{LabelFusion} produces annotated segmentation and pose data by first building a dense scene reconstruction of a trajectory around the object with \emph{ElasticFusion}, registering the model to the dense  point cloud, and transforming the object poses into each individual frame in the camera trajectory. We benchmark both \emph{SegICP} and \emph{SegICP-DSR} against  this $4493$ frame \emph{LabelFusion} dataset  using a ray-casting resolution of $250\,px$. 
	
	Our results are illustrated in Figure~\ref{fig:labelfusion-dataset} with histograms of the pose estimation results and the success measure indicated in red. \emph{SegICP-DSR} outperformed \emph{SegICP} considerably in both accuracy and variance. \emph{SegICP-DSR} had an average position error of $7.9\,mm, \sigma\mathord{=}7.6\,mm$, and an average angle error of $1.7\deg, \sigma\mathord{=}0.7\deg$, with the angle error defined as $\theta = cos^{-1}(2\langle q_1,q_2\rangle^2-1)$, while successfully identifying the pose in $96.7\%$ of images (with metric $<50 mm$ and $< 15\deg$ \cite{Wong2017segicp}).  \emph{SegICP-DSR} achieves notably higher success than \emph{SegICP} due to reconstructive segmentation, which leverages a time history of segmentations; whereas \emph{SegICP} is brittle to instantaneous false detections of the segmentor (\emph{SegNet}) and noise in the depth map. 
	
	For this dataset, \emph{SegICP-DSR} achieves $29\%$ increase in accuracy and $14\%$ increase in success as compared to \emph{SegICP}.

	
	\begin{figure}
		\subfigure{\includegraphics[width=1.25in]{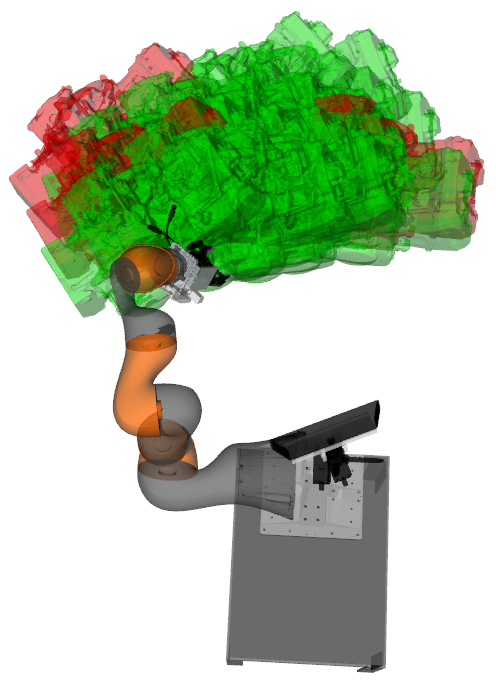}}%
		\vspace{-145pt}%
		\begin{table}[H]
			\hspace{97pt}%
			\begin{tabular}{|c|c|c|}
				\hline 
				\emph{SegICP} \cite{Wong2017segicp}& \cellcolor[HTML]{EFEFEF}\textbf{Avg. range} & \cellcolor[HTML]{EFEFEF}\textbf{Avg. std.} \\ \hline
				\cellcolor[HTML]{EFEFEF}$x$ & $5.6\,mm$              & $1.0\,mm$         \\ \hline
				\cellcolor[HTML]{EFEFEF}$y$ & $10.1\, mm$               &  $1.9\, mm$        \\ \hline
				\cellcolor[HTML]{EFEFEF}$z$ & $9.4\, mm$               &  $1.6\, mm$        \\ \hline
				\cellcolor[HTML]{EFEFEF}$\gamma$ (roll) & $2.73\, \deg$               &  $0.52\,\deg$        \\ \hline
				\cellcolor[HTML]{EFEFEF}$\beta$ (pitch )& $1.66\,\deg$               &  $0.35\,\deg$        \\ \hline
				\cellcolor[HTML]{EFEFEF}$\alpha$ (yaw)& $1.21\,\deg$               &  $0.25\,\deg$        \\ \hline
			\end{tabular}
		\end{table} 
		\vspace{-17.5pt}%
		\begin{table}[H]
			\hspace{95pt}%
			\begin{tabular}{|c|c|c|}
				\hline 
				\emph{SegICP-DSR}& \cellcolor[HTML]{EFEFEF}\textbf{Avg. range} & \cellcolor[HTML]{EFEFEF}\textbf{Avg. std.} \\ \hline
				\cellcolor[HTML]{EFEFEF}$x$ & $0.9\,mm$              & $0.2\, mm$         \\ \hline
				\cellcolor[HTML]{EFEFEF}$y$ & $1.8\, mm$               &  $0.5\, mm$        \\ \hline
				\cellcolor[HTML]{EFEFEF}$z$ & $1.0\, mm$               &  $0.3\, mm$        \\ \hline
				\cellcolor[HTML]{EFEFEF}$\gamma$ (roll)  & $0.19\,\deg$               &  $0.05\,\deg$        \\ \hline
				\cellcolor[HTML]{EFEFEF}$\beta$ (pitch )  & $0.21\,\deg$               &  $0.06\,\deg$        \\ \hline
				\cellcolor[HTML]{EFEFEF}$\alpha$ (yaw) & $0.19\,\deg$               &  $0.05\,\deg$        \\ \hline
			\end{tabular}
		\end{table} 
		\vspace{-10.5pt}
		\caption{\textbf{Experiment 2: ``Engine oil-filler cap grasping with the KUKA iiwa arm.''} $17/20$ engine poses resulted in successful grasps. Of the three that failed, two managed to grasp the fill port assembly (meaning that the grasp was fractions of a $cm$ too deep). Grasp results are shown (left) with green indicating successes and red corresponding to engine poses that failed. Among all these engine poses, \emph{SegICP-DSR} achieved a ${\sim}6\times$ smaller pose standard deviation as compared to \emph{SegICP} (right).}
		\label{fig:grasp-cap}
	\end{figure}

	\begin{figure*}
		\centering
		\begin{table}[H]
			\begin{tabular}{|c|c|c|c|c|c|c|}
				\hline 
				& \cellcolor[HTML]{EFEFEF}\textbf{Success} & \cellcolor[HTML]{EFEFEF}\textbf{Successful poses} &
				\cellcolor[HTML]{EFEFEF}\textbf{Euclidean std. (success)} &  
				\cellcolor[HTML]{EFEFEF}\textbf{Euler std. (success)}  &
				\cellcolor[HTML]{EFEFEF}\textbf{Euclidean std. (failure)} &
				\cellcolor[HTML]{EFEFEF}\textbf{Euler std. (failure)} \\ 
				\hline
				\cellcolor[HTML]{EFEFEF}\emph{SegICP} \cite{Wong2017segicp} & $27/50$        & $5/10$          & $1.67\,mm$      & $ 0.43 \deg$ &$5.05 \,mm$ &$1.68 \deg$  \\ \hline

				\cellcolor[HTML]{EFEFEF}\emph{SegICP-DSR} & $48/50$        & $10/10$        &  $0.11 \, mm$ & $0.02 \deg $&   $0.51 \, mm$ & $0.19 \deg$   \\ \hline
			\end{tabular}
		\end{table}
		\vspace{-12.5pt}
		\caption{\textbf{Experiment 3: tight-tolerance, semi-rigid insertions.} \emph{SegICP-DSR} achieves a considerably smaller standard deviation in its engine pose detections (overall euclidean std: $0.15\,mm$, euler std: $0.03\deg$) compared to \emph{SegICP} (overall euclidean std: $4.37\,mm$, euler std: $1.42\deg$) for each scene and as a result, achieves overall higher success rates for the insertion task. \emph{Successful poses} are the unique engine poses that resulted in at least $80\%$ ($4/5$) successful insertions.} 
		\label{fig:funnel-insert}
	\end{figure*}
	
	\vspace{5pt}
	\textbf{-- Experiment 2: Tight-tolerance grasping.} We demonstrate the accuracy of \emph{SegICP-DSR} by grasping and removing the oil filler cap from a motorcycle engine (see Fig. \ref{fig:exp-pic}). Using only our pose estimate, knowledge of the object geometry, and inverse kinematics, we compute a motion plan for this grasping task. We randomly placed the engine in arbitrary poses within reach of the robot, first running \emph{SegICP}, then \emph{SegICP-DSR} to acquire a time history of pose estimates for comparison. 
	As \emph{SegICP} did not provide a high enough quality pose for successful grasping, we only performed grasps with \emph{SegICP-DSR}. First, we directly use our pose estimate to derive the filler cap $6$-DOF pose and solve for a trajectory to grasp the cap using \emph{Drake}\footnote{Drake is a planning, control, and analysis toolbox for nonlinear dynamical systems \href{http://drake.mit.edu}{\texttt{http://drake.mit.edu}}.}. Of the twenty trials, the robot successfully performed $17/20$  grasps.  The trial engine poses are illustrated in Figure~\ref{fig:grasp-cap} (colored as green and red to indicate successes and failures, respectively), along with statistics for the  pose estimates from both \emph{SegICP} and \emph{SegICP-DSR}. \emph{SegICP-DSR} provided pose estimates which were ${\sim}6\times$ more accurate than \emph{SegICP} with an  average range (difference between max and min) for each component of the pose estimate ($x,y,z,\gamma,\beta,\alpha$) that was also similarly smaller for \emph{SegICP-DSR}. The range and standard deviation reported are over multiple observations for each unique engine pose. 
	
	\vspace{5pt}	
	\textbf{-- Experiment 3: Tight-tolerance, peg-in-hole insertion.} We categorize the accuracy of our pose estimation in the loop with a difficult, tight-tolerance insertion task. The task is to insert a plastic blue funnel (with a tip radius of $8.0\,mm$) into the engine fill port ($11.5\,mm$ radius), which places a sub-$cm$ tolerance on the pose-estimation error if the robot is to successfully execute an insertion without distorting the funnel or moving the engine. 
	
	We executed $100$ insertions on the KUKA iiwa arm by first estimating the engine's pose and finding a kinematic trajectory relative to the estimated object frame; these consisted of ten randomly positioned engine poses in reach of the robot. For each pose, we attempted to insert the funnel into the engine fill port five times each with \emph{SegICP} and \emph{SegICP-DSR}. The results are indicated in  Figure~\ref{fig:funnel-insert}. Again, the detected poses using \emph{SegICP-DSR} exhibited ${\sim}30x$ smaller standard deviation (\emph{SegICP-DSR} euclidean std: $0.15\,mm$, euler std: $0.03\deg$ compared to \emph{SegICP} euclidean std: $4.37\,mm$, euler std: $1.42\deg$), which resulted in higher success rates ($48/50$) for the insertion task compared to \emph{SegICP} ($27/50$). High variability in the pose estimation of \emph{SegICP} appears correlated to the trials that exhibited insertion failures, given by $>5.0 \,mm$ euclidean std. and $>1.58 \deg$ euler std. 
	
	\textbf{Success measure.} We categorized a successful insertion by the blue funnel's tip being completely inside the engine fill port ($2.1\,cm$ depth), see Figure~\ref{fig:exp-pic}. If the arm pushes the engine considerably due to poor pose estimation and the tip of the funnel collides with the edge of the fill port, it is categorized as a failure. \emph{Successful poses} are then the unique engine poses (out of the possible ten) that resulted in at least $80\%$ ($4/5$) successful insertions.

	%

	\section{Conclusion}
	We present a real-time, dense, semantic scene reconstruction and pose estimation system, \emph{SegICP-DSR}, that achieves $mm$-level pose accuracy and standard deviation ($7.9 \,mm, \sigma\mathord{=}7.6 \,mm$ and $1.7\deg, \sigma\mathord{=}0.7\deg$), enabling tight-tolerance manipulation in unstructured environments. We acquired these results on thousands of video frames and demonstrated that with accurate perception alone, we can solve for kinematic trajectories using \emph{SegICP-DSR} pose estimates to accomplish tight-tolerance manipulation tasks with high probabilities for success ($48/50$ successful tight-tolerance insertions). Because \emph{SegICP-DSR} operates over a fused, dense scene reconstruction, it produces a smoothed pose estimate between consecutive frames (overall euclidean std: $0.15\,mm$, euler std: $0.03\deg$) without additional filtering. 
	
	We are actively investigating more sophisticated registration methods (e.g. local point-to-mesh and global registration) to further increase our pose estimation accuracy. The availability of real-time, high quality pose estimation improves the applicability and utility of advanced task planners \cite{Kaelbling2013} and motion planners which incorporate dynamics \cite{plancher2017constrained,manchester2017dirtrel}. Accurate initial pose estimates can serve as a starting point for local feature tracking or learned controllers \cite{Mahler2017dexnet, Levine2016end} --- ultimately we are interested in a combination of these methods in order to solve more complex manipulation tasks in unstructured environments.

	\bibliographystyle{IEEEtran}
	\footnotesize{\bibliography{icp}}
\end{document}